\documentclass{llncs}
\usepackage{epsfig}
\usepackage{epsfig,latexsym, amssymb, amsmath}
\usepackage{lscape}
\usepackage{multirow}

\newcommand\T{\rule{0pt}{2.6ex}}

\newcommand{\hide}[1]{}

\title{Experimental Evaluation of Branching Schemes for the CSP}

\author{Thanasis Balafoutis\inst{1} \and Anastasia Paparrizou\inst{2} \and Kostas Stergiou\inst{2}} 

\institute{Department of Information and Communication Systems
  Engineering,\\ University of the Aegean, Greece.
  \and
Department of Informatics and Telecommunications Engineering,\\
  University of Western Macedonia, Greece.}

\begin{document}

\maketitle




\begin{abstract}

The search strategy of a CP solver is determined by the variable and value ordering heuristics it employs and by the branching scheme it follows.
Although the effects of variable and value ordering heuristics on search effort have been widely studied, the effects of different branching schemes have received less attention. 
In this paper we study this effect 
through an experimental evaluation that includes
standard branching schemes such as 2-way, $d$-way, and dichotomic domain splitting, as well as variations of set branching where branching is performed on sets of values. We also propose and evaluate a generic approach to set branching where the partition of a domain into sets is created using the scores assigned to values by a value ordering heuristic, and a clustering algorithm from machine learning. Experimental results 
demonstrate that although exponential differences between branching schemes, as predicted in theory between 2-way $d$-way branching, are not very common, still the choice of branching scheme can make quite a difference on certain classes of problems. Set branching methods are very competitive with 2-way branching and outperform it on some problem classes. A statistical analysis of the results reveals that our generic clustering-based set branching method is the best among the methods compared.


\end{abstract}

\section{Introduction}


Complete algorithms for CSPs are based on exhaustive backtracking search interleaved with constraint propagation. Search is typically guided by variable and value ordering heuristics and makes use of a specific branching scheme like 2-way or $d$-way branching. Although the impact of variable and value ordering heuristics on search performance are topics that have received very wide attention from the early days of CP, the impact of different branching schemes has not been as widely studied. As a result, the majority of modern finite domain CP solvers offer a wide range of variable and value ordering heuristics for the user/modeller to choose from, but at the same time they typically always employ 2-way branching. Some solvers allow for the user to implement different branching schemes, but it is not clear in which cases this may be desirable, and which particular scheme should be prefered.

In 2-way branching, after a variable $x$ with domain $\{a_1,\ldots, a_d\}$
is chosen, its values are assigned through a sequence of binary choices \cite{sabin97}. The first choice point creates two branches, corresponding to the assignment of $a_1$ to $x$ (left branch) and the removal of $a_1$ from the domain of $x$ (right branch).
An alternative branching scheme which was extensively used in the past, and is still used by some solvers, is $d$-way branching. In this case, after variable $x$ is selected, $d$ branches are built, each one corresponding to one of the $d$ possible value assignments of $x$.  2-way branching was described by Freuder and Sabin within the MAC algorithm \cite{sabin97} and in theory it
can achieve exponential savings in search effort compared to $d$-way branching \cite{mitchell05}. However, the few experimental studies comparing 2-way and $d$-way branching have not displayed significant differences between the two \cite{park04,smith05}. Very recently Balafoutis and Stergiou showed that depending on the variable ordering heuristic used there can be from marginal to exponential differences between the two schemes \cite{balafoutis2010}.

Another technique that is sometimes used is dichotomic domain splitting \cite{dincbas88}. This method originates from numerical CSPs and proceeds by splitting the current domain of the selected variable into two sets, usually based on the lexicographical ordering of the values. In this way branching is performed on the two created sets and the branching factor is reduced to two.
Although domain splitting drastically reduces the branching factor, it can result in a much deeper search tree since the effects of propagation after a branching decision may be diminished.

In addition to these standard schemes, techniques that group together the values of the selected variable, and branch on these created groups instead of individual values, have been proposed \cite{haselbock93,larrosa97,silaghi99,beckwith01,vanHoeve04,kitching09}. The criteria used for the grouping of values and the methods used to perform the grouping can be different, but all these techniques aim at reducing the size of the search tree. In this paper, following \cite{kitching09}, we call any such method a {\em set branching} method.

Our first goal in this paper is to experimentally study the effect of different branching schemes for finite domain CSPs on search performance. Although some existing branching methods have been compared to one another (e.g. \cite{smith05}), to our knowledge this is the first systematic evaluation of several existing alternatives. 
In addition, we propose and study a generic set branching method where the partition of a domain into sets is created using the scores assigned to values by a value ordering heuristic, and a clustering algorithm. Before employing such a method, two fundamental questions need to be adressed: What is the measure of similarity between values, and how do we partition domains using such a measure?
Most of the approaches to set branching that have been proposed in the past have either used very strict measures of similarity or are problem specific. Our method offers a generic solution to both the problem of similarity evaluation and the partitioning of domains. For the former we exploit the information acquired from the value ordering heuristic, while for the latter we use a clustering algorithm from machine learning.


Experimental results from a wide range of benchmarks demonstrate that exponential differences between branching schemes, as predicted in theory between 2-way $d$-way, are not very common. But although the choice of branching scheme does not have as a profound effect as the choice of variable ordering heuristic, it can still make a difference. The generic set branching methods we evaluate outperform the standard 2-way branching scheme in many problem classes resulting in better average performance. It is notable that our clustering-based set branching method displays very promising results without any tuning of the clustering algorithm applied. Importantly, a statistical analysis of the experimental results reveals that this method is the best among the methods compared.

The rest of the paper is organized as follows. Section~\ref{section-background} gives necessary background. In Section~\ref{section-set} we discuss past work on set branching for CSPs and propose a new generic method for set branching. In Section~\ref{section-experiments} we report results from an experimental evaluation of the various branching schemes including a statistical analysis. Finally, in Section~\ref{section-conclusions} we conclude.


\section{Background}

\label{section-background}


A \emph{Constraint Satisfaction Problem} (CSP) is a tuple
(\emph{X, D, C}), where \emph{X} is a set containing \emph{n}
variables \{\emph{$x_1, x_2,..., x_n$}\}; \emph{D} is a set of
domains \{\emph{$D(x_1)$, $D(x_2)$,..., $D(x_n)$}\} for those
variables, with each $D(x_i)$ consisting of the possible values
which $x_i$ may take; and \emph{C} is a set of constraints
\{\emph{$c_1, c_2,..., c_e$}\}
between variables in subsets of
\emph{X}. Each constraint $c \in C$ expresses a relation $rel(c)$ defining the variable assignment combinations that are allowed for the variables in
the scope of the constraint \emph{vars($c$)}.


Complete search algorithms for CSPs are typically based on backtracking depth-first search where branching decisions (e.g. variable assignments) are interleaved with constraint propagation. 
Search is guided by variable ordering heuristics and value ordering heuristics. 

One of the most efficient general purpose variable ordering heuristics that have been proposed is {\em dom/wdeg} \cite{bhls04}. This heuristic assigns a weight to each constraint, initially set to one. Each time a constraint causes a conflict, i.e. a domain wipeout, its weight is incremented by one. Each variable is associated with a {\em weighted degree}, which is the sum of the weights over all constraints involving the variable and at least another unassigned variable. The {\em dom/wdeg} heuristic chooses the variable with minimum ratio of current domain size to weighted degree. 


A well-known generic value ordering heuristic for binary CSPs is Geelen's {\em promise} \cite{geelen92}. For each value $a\in D(x)$ this heuristic counts the number of values that are compatible with $a$ in each future unassigned variable that $x$ is constrained with. The product of these counts is the promise of $a$. The value with the maximum promise is selected.

\section{Branching Schemes}

\label{section-set}

\begin{figure}[htb]
\begin{tabular}{c}
\includegraphics{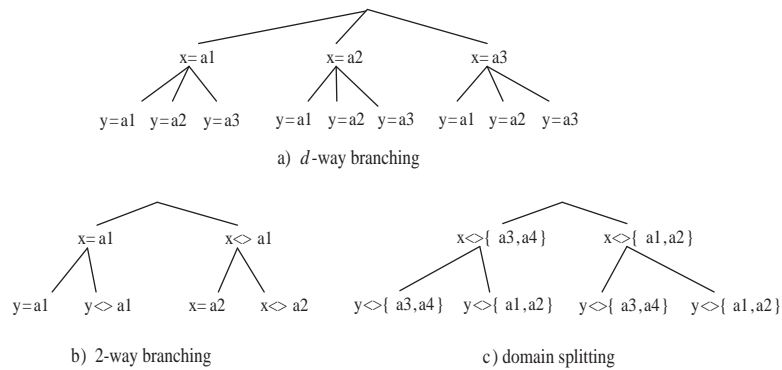}
\end{tabular}
\vspace{-5mm}
\caption{Examples of search trees for the three branching schemes.}
\label{fig-branching}
\end{figure}

From the early days of CSP research, search algorithms were usually implemented using either a \emph{d-way} or a \emph{2-way} branching scheme. The former works as follows. After a variable $x$ with domain $D(x)=\{a_1, a_2,..., a_d\}$ is selected, $d$ branches are created, each one corresponding to a value assignment of $x$. In the first branch, value $a_1$ is assigned to $x$ and constraint propagation is triggered. If this branch fails, $a_1$ is removed from $D(x)$. Then the assignment of $a_2$ to $x$ is made (second branch), and so on. If all $d$ branches fail then the algorithm backtracks. An example of a search tree explored with $d$-way branching is shown in Figure~\ref{fig-branching}a.

In \emph{2-way} branching, after a variable $x$ and a value $a_i \in D(x)$ are selected, two branches are created. In the left branch $a_i$ is assigned to $x$, or in other words the constraint $x$=$a_i$ is added to the problem and is propagated. In the right branch the constraint $x\neq a_i$ is added to the problem and is propagated. If there is no failure then any variable can be selected next (not necessarily $x$). If both branches fail then the algorithm backtracks. Figure~\ref{fig-branching}b shows a search tree explored with 2-way branching.

There are two differences between these branching schemes.
In 2-way branching, if the branch assigning a value $a_i$ to a variable $x$ fails then the removal of $a_i$ from $D(x)$ is 
propagated. Instead, $d$-way branching tries the next available
value $a_j$ of $D(x)$. Note that the propagation of $a_j$ subsumes the propagation of $a_i$'s removal.
In 2-way branching, after a failed branch corresponding to an assignment $x$=$a_i$, and assuming the removal of $a_i$ from $D(x)$ is then propagated successfully, the algorithm can choose to branch on any variable (not necessarily $x$), according to the variable ordering heuristic.
In $d$-way branching the algorithm has to again branch on $x$ after $x$=$a_i$ fails.

Another option, that originates from numerical CSPs, is dichotomic {\em domain splitting} \cite{dincbas88}. This method proceeds by splitting the current domain of the selected variable into two sets, usually based on the lexicographical ordering of the values.  Once the domain has been split, the second set of values is removed from the domain and this removal is propagated. In this way branching is performed on the two created sets and the branching factor is reduced to two. However, domain splitting tends to achieve weaker propagation compared to $d$-way and 2-way branching. So, although it drastically reduces the branching factor, it can result in a much deeper search tree. Domain splitting is mostly used on optimization problems and especially when the domains of the variables are very large.
An example of a search tree explored with domain splitting is shown in Figure~\ref{fig-branching}c.

\subsection{Set Branching}


Very recently, Kitching and Bacchus explored the applicability of {\em set branching} for constraint optimization problems \cite{kitching09}. The basic idea is to group together values that offer similar improvement to the currently computed bounds. In this way entire groups of values that offer no improvement to the bounds can be refuted, resulting in smaller tree sizes.

In this paper we use the term {\em set branching} to refer to any branching technique that, using some similarity criterion, identifies values that can be grouped together and branched on as a set. Dichotomic domain splitting and 2-way branching can be seen as manifestations of this generic method that use simple grouping criteria. Domain splitting creates two sets of values based on their lexicographical ordering. 2-way branching splits the domain into two sets where the first includes a single value and the second the rest of the values. In general, in order to define a set branching technique, two questions need to be addressed: {\em What is the measure of similarity between values, and how are domains partitioned using such a measure}?

The idea of set branching for CSPs has been explored in the past. Freuder introduced the notion of interchangeability, substitutability, and their weaker, but tractable, neighborhood versions as means to identify values with similar behavior \cite{freuder91}. Two values of a variable are neighborhhood interchangeable iff they have exactly the same supports in all constraints. One value $a$ is neighborhood substitutable for another value $b$ if the set of values inconsistent wth $a$ is a subset of the values inconsistent with $b$.
These notions were exploited, for example in \cite{haselbock93,beckwith01,prestwich04}, to group together values when branching and in this way perform set branching. The drawback of these techniques is that their conditions are too strong, as in many problems neighborhood interchangeable and substitutable values are very rare.

Larrosa investigated the merging of similar subproblems during search using forward checking \cite{larrosa97}. According to this approach, values whose assignment leads to similar subproblems are grouped together and branched on as a set. Experiments performed on crossword puzzle generation problems displayed promising results. However, the measure of subproblem similarity and the algorithm used to partition the domains according to this measure are both problem specific.

Silaghi et al. proposed a method for partitioning the domains of variables based on the Cartesian product representation of the search space \cite{silaghi99}. This method is particularly suitable for finding all solutions but it requires an explicit extensional representation of the constraints in the problem.

A generic and simple approach to set branching that can be applied on a wide range of problems was proposed by van Hoeve and Milano \cite{vanHoeve04}.
In this approach, values that are ``tied'' according to their value ordering heuristic score are grouped together and branching is performed on the sets of values created. Assignment of specific values to variables is postponed until lower levels of the search tree (which is also done in Larossa's method).
Experiments using both depth-first search and limited discrepancy search displayed promising results. However, this method relies heavily on the particular value ordering heuristic used and the number of ties produced by the value ordering heuristic, which can be quite low in many cases. Also, this method distinguishes between values that have very close but not equal scores and as a result such values will be placed into different sets.
As noted in \cite{vanHoeve04}, the concept of a tie can be extended to refer to values having close scores. In this paper we explore this idea further.


\subsection{Clustering for Set Branching}


As we intend to apply set branching dynamically throughout search, after selecting a variable $x$ with current domain $D(x)=\{a_1,\ldots,a_d\}$, we are faced with the following problem. We have to create a partition $S_{D(x)}=\{s_1,\ldots,s_m\}$ of $D(x)$ into $m$ sets s.t. each value $a_i\in D(x)$ belongs to only one set $s_j\in S$. Ideally, we want all the values that have been assigned to a specific set to be similar according to some measure of similarity.

Following van Hoeve and Milano, we use a generic measure of similarity that is based on the score of the values according to a value ordering heuristic.
In order to perform the dynamic partitioning of domains into sets, we propose the use of clustering algorithms from machine learning. Our approach can be summarized as follows. A value ordering heuristic is used to assign a score $v_i$ to each value $a_i\in D(x)$. The collection of  $d$ items (values) and the matrix of their scores are given as input to a clustering algorithm. The output of the algorithm will be the partition $S_{D(x)}=\{s_1,\ldots,s_m\}$.

Compared to \cite{vanHoeve04} our approach has the following potential benefits. First, not only will tied values be placed in the same set, but with high probability so will values that have very close scores. Hence, there will be fewer sets, resulting in lower branching factor. Second, in cases where there are no ties, the method of \cite{vanHoeve04} uses $d$-way branching. In contrast, our approach will still partition the domain if there are groups of values with similar score.

The algorithm we currently use to create the clustering of values is {\em x-means} \cite{pelleg00}. This is an extension of the well known k-means algorithm that is considerably faster and does not require to predetermine the desired number of clusters. The algorithm starts with randomly selected points (values in our case) as cluster centroids and iteratively improves the computed clustering until a fixpoint is reached. Several parameters of the algorithm can be tuned to give more accurate results on a specific application, including the starting centroids, the number of iterations, the measure of distance between points, etc. Although we intend to investigate this in the future, in the experiments reported below we use the Weka implementation of the x-means algorithm as is, without any tuning.


\hide{

\begin{description}

\item[heuristic\_score] A value ordering heuristic is used to assign a score to each value $a_i\in D(x)$. Values are then clustered into sets according to their heuristic score.

\item[support\_similarity] For each value $a_i\in D(x)$ we identify its supports in all constraints where $x$ participates. The values are then clustered according to the similarity of their supports.

\item[impact\_similarity] For each value $a_i\in D(x)$ we temporarily assign $x$ to $a_i$, propagate the assignment, and record the resulting domains of the future variables. The values are then clustered according to the similarity of the domains.

\end{description}

}

\section{Experimental Evaluation}

\label{section-experiments}


We have experimented with 350 instances from ten classes of real world, academic, patterned, and random CSPs taken from C.Lecoutre's XCSP repository. We included both satisfiable and
unsatisfiable instances. Each selected instance involves constraints defined either
in intension or in extension. The CSP solver used in our experiments is a generic solver and has been
implemented in the Java programming language. This solver
essentially implements the M(G)AC search algorithm, where (G)AC-3
is used for applying (G)AC. Since our solver does not yet support
global constraints (apart from the table constraint) , we have left experiments with problems that include such constraints as future work. All experiments were run on an Intel dual core PC T4200 2GHz with 3GB RAM.

For a fair evaluation of the different branching schemes we use the same propagation method during search (arc consistency), the same variable ordering heuristic ({\em dom/wdeg} \cite{bhls04}) and value ordering heuristic ({\em Geelen's promise} \cite{geelen92}). The promise metric is calculated over all the visited nodes of the search tree. This penalizes run times and as a result may be inefficient in some problems, but for the purposes of this initial investigation we only wanted to use a reasonably sophisticated value ordering heuristic throughout all the tried instances. In the future we intend to experiment with different value ordering heuristics and study their effect on the performance of the clustering set branching method.

We compare the following branching schemes:

\begin{description}

\item[2-way] Values are chosen in descending order of their promise.

\item[$d$-way] Values are chosen in descending order of their promise.

\item[domain splitting] The values are ordered according to their promise and then the domain is split in half. The part with the top ranked values is tried first.

\item[ties set branching] This is the method of \cite{vanHoeve04} where values with the same promise form a set. The sets are tried in descending order of promise. 

\item[clustering set branching] This is our method where x-means is used to partition the domain into sets based on the promise of the values. The sets are tried in descending order of promise. Note that the clusters are linearly ordered since clustering is done over only one dimension.  

\end{description}

The two set branching methods have been implemented using a 2-way and a $d$-way branching style, giving four alternatives. More specifically, past works on set branching for CSPs perform set branching using a $d$-way style. That is, once the partition of the domain $S_{D(x)}=\{s_1,\ldots,s_m\}$ is created, search proceeds by removing from $D(x)$ any value $a$, s.t. $a \notin s_1$, and propagating. If there is a failure, the same process is repeated for $s_2$ and so on.
We have also implemented and evaluated 2-way style set branching. In this case the generated sets are tried in a series of binary choices. That is, after the reduction of $D(x)$ to $s_1$ fails, we propagate the removal from $D(x)$ of all the values in $s_1$. If this succeeds then we reduce $D(x)$ to $s_2$ and so on.

We must clarify here that in all the ``2-way style'' branching variants (domain splitting, ties, clustering) the set branching method allows to jump from one variable to another as standard 2-way branching does.

In addition, for domain splitting and the set branching methods we have tried two options: 1) Domain splitting (resp. set branching) is performed throughout search on all variables. 2) Domain splitting (resp. set branching) is performed on a variable only if its domain size is greater than a certain percentage of its original domain size. 
We have tried several values for this percentage, with 25$\%$ giving the best results. This can improve the performance of domain splitting by 30$\%$ on average, and it can offer (minor) improvement to set branching. Therefore, in the reported experiments with these methods Option 2 is followed.


\hide{

\begin{figure}
  \begin{tabular}{cc}
    \includegraphics[height=1.6in]{dichotomicBars.eps} &
    \includegraphics[height=1.6in]{clusterBars.eps}\\
    (a)&(b)
  \end{tabular}
\caption{Visited nodes percentage reduction, relative to domain splitting when it is performed throughout search on all variables (a) for the domain splitting method (b) for the d-way clustering set branching method.}
\label{fig:bars}
\end{figure}

}

\begin{table}[htb]
\caption{Cpu times (t), and nodes (n) from specific instances. Cpu times are in seconds. The best result for each instance is given in bold.}
\centering
\begin{scriptsize}
\begin{tabular}{|c|c|c|c|c|c|c|c|c|c|}
\hline
 \T& &  & & & $d$-$way$ & $2$-$way$ & $d$-$way$ & $2$-$way$\\
$Problem$ \T  & & $d$-$way$ & $2$-$way$ & $dom$ & $ties$ & $ties$ & $clust.$ & $clust.$\\
$Class$ \T& & & & $split.$ & $set$ $branch.$ & $set$ $branch.$& $set$ $branch.$& $set$ $branch.$\\
\hline 
$frb35$-$17$-$2$ \T & t &\textbf{43.3} & 98.4 & 954 & 60.1 & 98.3 & 134 & 154\\
$(sat)$ \T & n & 16241 & 45098 & 515909 & 27160 & 50713 & 58633 & 75743\\ \hline
$scen3$-$f11$ \T & t & 73.7 & \textbf{6.9} & 33.8 & 40.1 & 11.3 & 43.5 & 14.5\\
$(unsat)$ \T & n & 11056 & 1739 & 5318 & 11019 & 4021 & 13631 & 5705\\ \hline
$pigeons$-$30$-$ord$ \T & t & 2435 & \textbf{572} & 762 & 1259 & 773 & 1322 & 639\\
$(unsat)$ \T & n & 376384 & 135031 & 128286 & 338049 & 247792 & 364343 & 228190\\ \hline
$geo50$-$20$-$d4$-$75$-$7$ \T & t & 472 & 1338 & 2815 & \textbf{190} & 1309 & 365 & 543\\
$(sat)$ \T & n & 108027 & 404918 & 686333 & 58411 & 443724 & 111505 & 174716\\ \hline
$langford$-$2$-$10$ \T & t & 300 & 129 & 605 & \textbf{108} & 120 & 116 & 127\\
$(unsat)$ \T & n & 199104 & 247286 & 372733 & 199609 & 235912 & 203580 & 238314\\ \hline
$driverw$-$09$ \T& t & 177 & 145 & 243 & \textbf{103} & 164 & 180 & 143\\
$(sat)$ \T & n & 75625 & 93236 & 97180 & 46823 & 76510 & 77509 & 64798\\ \hline
$qcp$-$15$-$120$-$6$ \T & t & 23.8 & 12.4 & 26 & 28.8 & \textbf{9.6} & 133 & 94.6\\
$(sat)$ \T & n & 19074 & 20179 & 19353 & 33003 & 12019 & 136599 & 99847\\ \hline
$qcp$-$15$-$120$-$8$ \T& t & 50 & 35.4 & 53.2 & 44.4 & 130 & \textbf{1.01} & \textbf{1.01}\\
$(sat)$ \T & n & 38227 & 49680 & 38551 & 46188 & 146342 & 845 & 845\\ \hline
$geo50$-$20$-$d4$-$75$-$11$ \T& t &  41.6 & 38.9 & 94.2 & 32.5 & 37.9 & \textbf{12.2} & 15.1\\
$(sat)$ \T & n & 9027 & 10044 & 21926 & 8990 & 12620 & 3486 & 5111\\ \hline
$queensKnights$-$15$-$5$-$add$ \T& t &  1506 & 1001 & 2245 & 1502 & 737 & 999 & \textbf{594}\\
$(unsat)$ \T & n & 42154 & 15393 & 86199 & 42309 & 38836 & 28312 & 30890\\ \hline
\end{tabular}
\end{scriptsize}
\label{table:extremeResults} 
\end{table}

Table~\ref{table:extremeResults} compares the various branching methods on specific instances from the tested problem classes. We display CPU times as well as nodes. A node in 2-way branching can correspond to a value assignment or to a value removal, while in d-way branching it can only correspond to a value assignment. Hence, they cannot be compared directly. 
The instances in this table are chosen to highlight the gaps in performance that can occur when using different branching schemes. As can be seen any method can be the best on a given instance, and there can be very considerable variance in the performance of the methods.
For instance, clustering set branching can be very effective on certain problems (e.g. qcp-15-120-8) but it can also be quite ineffective on others (e.g. qcp-15-120-6). However, these are some of the most `extreme' instances. Exponential differences, as predicted between 2-way and $d$-way in theory, occured rarely\footnote{But this observation concerns the variable ordering heuristic and propagation method used here and may not generalize as shown in \cite{balafoutis2010}.}.

\begin{table}[htb]
\caption{Average speed-up (positive values) or slow-down (negative values) achieved by 2-way branching compared to the other branching methods. Cpu time (t) in seconds and visited nodes (n) have been measured.}
\centering
\begin{scriptsize}
\begin{tabular}{|c|c|c|c|c|c|c|c|c|}
\hline
 \T& $\%$ & & &  & $d$-$way$ & $2$-$way$ & $d$-$way$ & $2$-$way$\\
$Problem$ \T&  $graph$ & & $d$-$way$ & $dom$ & $ties$ & $ties$ & $clust.$ & $clust.$\\
$Class$ \T& $density$ & & & $split.$ & $set$ $branch.$ & $set$ $branch.$& $set$ $branch.$& $set$ $branch.$\\
\hline 
$langford$ \T& 1.045 & t & 2.88 & 5.08 & -1.21 & -1.11 & -1.20 & -1.04\\
$(unsat)$ \T&  & n & -1.27 & 1.52 & -1.26 & -1.06 & -1.23 & -1.03\\ \hline
$pigeons$ \T& 1 & t & 1.13 & 1.24 & -1.53 & -1.89 & -1.07 & -1.32\\
 $(unsat)$ \T&  & n & -1.21 & 1.33 & -1.7 & -1.66 & -1.25 & -1.12\\ \hline
$queensKnights$ \T&  0.70 & t & 1.49 & 1.99 & 1.75 & -1.21 & -1.02 & -1.48\\
 $(unsat)$ \T&  & n & 2.85 & 4.96 & 3.47 & 3.04 & 1.87 & 2.39\\ \hline
$forced$ $random$ \T & 0.65 & t & -1.22 & 1.88 & -1.30 & -1.03 & -1.14 & 1.14\\
 $(sat)$ \T& & n & -1.41 & 1.52 & -1.11 & -1.1 & -1.24 & 1.07\\ \hline
$geometric$ \T& 0.35 & t & -2.48 & 2.07 & -4.55 & -1.03 & -3.83 & -2.58\\
 $(sat)$ \T& & n & -3.02 & 1.79 & -3.77 & 1.18 & -3.53 & -2.25\\ \hline
$qcp-qwh$ \T& 0.125 & t & 1.78 & 2.34 & 1.28 & 1.99 & 6.08 & 5.63\\
 $(sat)$ \T& & n & -1.09 & 1.12 & -1.06 & 1.5 & 4.08 & 3.84\\ \hline
$driver$\T& 0.082 & t & 1.18 & 1.53 & -1.33 & 1.10 & 1.21 & 1.00\\
 $(sat)$ \T& & n & -1.23 & -1.06 & -1.71 & -1.24 & -1.23 & -1.43\\ \hline
$rlfap$ $(ScensMod)$\T & 0.052 & t & 5.39 & 3.07 & 3.52 & 1.07 & 3.70 & 1.26\\
 $(mixed)$ \T& & n & 4.63 & 2.73 & 4.28 & 1.77 & 4.94 & 2.1\\ \hline
$graphColoring$\T& 0.05 & t & -1.50 & 1.01 & -1.58 & 1.00 & -1.49 & -1.03 \\
 $(mixed)$ \T& & n & -1.28 & 1.15 & -1.18 & 1.14 & -1.17 & -0.92\\ \hline
\end{tabular}
\end{scriptsize}
\label{table:percentages1} 
\end{table}

In Tables~\ref{table:percentages1} and~\ref{table:percentages2} we summarize the results of our experimental evaluation. 
We use 2-way branching as the standard all other branching methods are compared against. In Table \ref{table:percentages1} we give the average slow-down (or speed-up) of the methods compared to 2-way for each problem class (the two quasigroup classes qcp and qwh are grouped together). We have mostly selected problem classes that contain either only satisfiable or only unsatisfiable instances. However, we  have also experimented with ``mixed'' problem classes. That is classes that contain both satisfiable and unsatisfiable instances. For example, on langford problems all instances are unsatisfiable and 2-way is 2.88 times better than $d$-way on average, while it is 1.2 times worse than $d$-way clustering set branching. As mentioned above, it is difficult to accurately compare the numbers of visited nodes under different branching schemes. However, in most problem classes the differences in Cpu times roughly reflect the differences in visited nodes. 

In Table \ref{table:percentages2} we give the percentage of instances, over all the tried instances, where each method was faster ($>1$), at least 2 times faster ($>2$), and at least 3 times faster ($>3$) than 2-way branching. Similarly for instances where each method was slower by $<1$, $<2$, and $<3$ times compared to 2-way.

Table \ref{table:percentages1} shows that although differences between methods can be quite large on single instances, the average differences between the most competitive methods are smaller. Dichotomic domain splitting is apparently the worst among the branching methods. However, it may fare better in problems with very large domain sizes\footnote{Most domains included between 2 and 50 values, with maximum 225.}. Excluding domain splitting, the other methods are usually no more that 2 times better or worse than 2-way branching on average. But there are cases where even the average differences are quite large.

The set branching methods, and especially the $d$-way style ones, have slightly better or very close performance compared to 2-way branching on most classes. Also, these methods clearly outperform $d$-way branching. Interestingly, the set clustering methods are typically very competitive on the denser classes.

\begin{table}[htb]
\caption{$\%$ categorization of all tried instances according to the performance of the branching methods compared to 2-way branching.}
\centering
\begin{scriptsize}
\begin{tabular}{|c|c||c|c|c|c|c|c|}
\hline
 \T& &  &   & $d$-$way$ & $2$-$way$ & $d$-$way$ & $2$-$way$\\
$Problem$ \T&  $speedup$  & $d$-$way$ & $dom$ & $ties$ & $ties$ & $clust.$ & $clust.$\\
$Class$ \T&   &  & $split$ & $set$ $branch.$ & $set$ $branch.$& $set$ $branch.$& $set$ $branch.$\\\hline
\multirow{6}{*}{all instances}
      & $>$1 & 29\% & 11\% & 47\% & 68\% & 50\% & 45\%\\
      & $>$2 & 8\% & 0\% & 8\% & 2\% & 15\% & 16\%\\
      & $>$3 & 2\% & 0\% & 3\% & 0\% & 10\% & 6\%\\
      & $<$1  & 71\% & 89\% & 53\% & 32\% & 50\% & 55\%\\
      & $<$2 & 24\% & 56\% & 21\% & 2\% & 21\% & 15\%\\
      & $<$3 & 11\% & 34\% & 6\% & 3\% & 11\% & 6\%\\
\hline
\end{tabular}
\end{scriptsize}
\label{table:percentages2} 
\end{table}

\hide{

Figure~\ref{fig:tiesVSclusters} compares the 2-way variants of \emph{ties set branching} and \emph{clustering set branching}. Each point in this plot corresponds to a single instance from the benchmarks. The $y$ and $x$ axes give cpu times for \emph{2-way clustering set branching} and \emph{2-way ties set branching} respectively. Therefore, a point below line $y=x$ represents an instance where  \emph{2-way clustering set branching} was faster. Both axes are logarithmic.

\begin{figure}
\label{fig:tiesVSclusters}
  \caption{A comparison of run times for the 2-way ties and 2-way clustering set branching schemes.}
  \begin{center}
    \includegraphics[height=2.2in]{ties_vs_clusters.eps}
  \end{center}
\end{figure}

As it can be clearly seen both branching schemes have on average a relatively close performance.

}

Table \ref{table:percentages2} shows that 2-way ties set branching is better than 2-way on most instances. However, the margins are usually small. This is because the number of ties that occur during search is usually low, meaning that 2-way ties set branching often emulates the standard 2-way scheme. The other set branching methods are better than 2-way on roughly half of the instances. However, they can be significantly better, and worse, on quite a few.


\begin{table}
\caption{Paired t-test measurements for evaluation of the significance of the experimental results. 2-way branching is compared with the other branching schemes.}
\centering
\begin{scriptsize}
\begin{tabular}{|c|c|c|c|c|c|}
\hline
  \T& Mean & SD & t-value  & 95\% C.I.\\ \hline
d-way  \T& -29.8 & 341.7 & -0.68 & (-116, 57) \\ \hline
domain splitting  \T& -241 & 456 & -4.1 & (-357, -125)\\ \hline
d-way ties set branching  \T& 9.48 & 326.3 & 0.23 & (-73.3, 92.3)\\ \hline
2-way ties set branching   \T& 31.7 & 234 & 1.06 & (-27.7, 91.1) \\ \hline
d-way clustering set branching  \T& 13.75 & 217.9 & 0.49 & (-41.6, 69)\\ \hline
2-way clustering set branching   \T& 32.4 & 182.5 & 1.4 & (-13.9, 78.7) \\ \hline
\end{tabular}
\end{scriptsize}
\label{table:ttest} 
\end{table}

In order to obtain a global view and to evaluate the statistical significance of
our experimental results, a set of paired t-tests were performed. In these tests we compared the CPU performance
of the 2-way branching scheme against all the other branching schemes, over all the instances used in the experiments. We measured the mean difference, standard deviation, t-value and the 95\% confidence
interval. The risk level (called alpha level) was set to 0.05.
Results are collected in Table \ref{table:ttest}.

As the results show, d-way branching and domain splitting are clearly inefficient compared to 2-way branching.
The mean CPU reduction in the all set branching techniques is always greater than zero with 2-way clustering set branching being slightly better. However, the negative values at the confidence interval indicate that this reduction was
not observed in all the tried instances. Although 2-way ties and clustering set branching achieve equivalent mean CPU reduction, the t-values score show that the spread (or variability) of the scores for 2-way clustering set branching is significantly higher compared to 2-way ties set branching. The t-value scores lead us to conclude that 2-way clustering set branching is a promising branching technique, since in our experiments it has displayed the best overall performance.

Finally, we have to mention that the number of clusters produced by x-means during search was usually quite low (2-3). In some cases, typically for small domain sizes, there was only one cluster generated because all values had similar score. In such a case our method switched to either d-way or 2-way branching depending on the style of set branching employed.

\section{Conclusions}


\label{section-conclusions}

We performed an experimental evaluation of branching methods for CSPs including the commonly used 2-way and $d$-way schemes as well as other less widely used ones. We also proposed and evaluated a generic set branching method that partitions domains into sets of values by using information provided by the value ordering heuristic as input to a 
clustering algorithm. Results showed that set branching methods, including our approach, are competitive and often better compared to standard 2-way branching. We now plan to investigate ways to achieve more efficient domain partitions by automatically tuning the parameters of the clustering algorithm. Also, it would be interesting to study clustering of domains using information from multiple value ordering heuristics.


\bibliography{extra}

\end{document}